\documentclass[11pt]{article} 

\usepackage[T1]{fontenc}
\usepackage{color}
\usepackage{graphicx}
\usepackage{hyperref}
\usepackage{lineno}
\usepackage{etoolbox} 

\usepackage{cite} 

\begin{document}


\newtoggle{arxiv}
\toggletrue{arxiv}

\title{Compression and the origins of Zipf's law for word frequencies}

\author{R. Ferrer-i-Cancho$^*$ \\
{\small $^1$Complexity \& Quantitative Linguistics Lab, LARCA Research Group,} \\
{\small Departament de Ci\`encies de la Computaci\'o.} \\ 
{\small Universitat Polit\`ecnica de Catalunya,} \\ 
{\small Campus Nord, Edifici Omega Jordi Girona Salgado 1-3.} \\ 
~\\
\iftoggle{arxiv}{ {\small (*) Send correspondence to rferrericancho@cs.upc.edu}
} {
{\small (*) Send correspondence to rferrericancho@cs.upc.edu. Phone: +34 934134028}
}
}

\date{}

\maketitle

\begin{abstract}
Here we sketch a new derivation of Zipf's law for word frequencies based on optimal coding. The structure of the derivation is reminiscent of Mandelbrot's random typing model but it has multiple advantages over random typing: (1) it starts from realistic cognitive pressures (2) it does not require fine tuning of parameters and (3) it sheds light on the origins of other statistical laws of language and thus can lead to a compact theory of linguistic laws. Our findings suggest that the recurrence of Zipf's law in human languages could originate from pressure for easy and fast communication.
\end{abstract}



\iftoggle{arxiv}{
} {
{\bf Short communication} ~\\

{\bf 6 pages, 0 figures and 0 tables. } \\
}
	
Zipf's law for word frequencies states that the probability of the $i$-th most frequent word obeys
\begin{equation}
p_i \approx i^{-\alpha}
\label{Zipfs_law_equation}
\end{equation}
with $\alpha \approx 1$ \cite{Alday2016a,Moreno2016a}. 
Here we explore the possibility that Zipf's law is a consequence of compression, the minimization of the mean length of the words of a vocabulary \cite{Ferrer2012d}. 
This principle has already been used to explain the origins of other linguistic laws: Zipf's law of abbreviation, namely, the tendency of more frequent words to be shorter \cite{Ferrer2012d,Ferrer2015a}, and Menzerath's law, the tendency of a larger linguistic construct to be made of smaller components \cite{Gustison2016a}. The mean length of words is reminiscent of the {\em minimum equation}, the cost function that Zipf put forward to shed light on the origins of Zipf's law of abbreviation: the mean length of words can be interpreted as a particular case of the minimum equation in which the energetic cost of a word is fully determined by its length \cite{Ferrer2012d}.
Our argument for the origins of Eq. \ref{Zipfs_law_equation} combines two constraints for compression that are ideal for natural languages: (1) non-singular coding, i.e. any two different word types should not be represented by the same string of letters or phonemes, and (2) unique decipherability, i.e. given a continuous sequence of letters or phonemes, there should be only one way of segmenting it into words \cite{Cover2006a}. The former is needed to reduce the cost of retrieving the original meaning. The latter is required to reduce the cost of determining word boundaries. Thus both constraints on compression and compression itself, are realistic cognitive pressures that are vital to fight against the now-or-never bottleneck of linguistic processing \cite{Christiansen2015a}.
 
Suppose that words are coded using an alphabet of $N$ letters (or phonemes) with $N > 1$ and that 
$p_i$ and $l_i$ are, respectively, the probability and the length of the $i$-th most probable word. Let us consider the optimal assignment of strings to different words assuming that words are made of at least one letter (or phoneme) and that the $p_i$'s are given (the $l_i$'s are determined by the string that is eventually assigned to every different word). On the one hand, optimal uniquely decipherable coding gives \cite{Cover2006a}
\begin{equation}
l_i = \lceil -\log_N p_i \rceil, 
\end{equation}
where $\lceil .. \rceil$ is the ceiling function. 
Thus 
\begin{equation}
l_i \approx -\log_N p_i. 
\label{optimal_coding_equation}
\end{equation}
On the other hand, optimal non-singular coding gives \cite{Ferrer2015a} 
\begin{equation}
l_i = \left\lceil \log_N \left(\frac{N-1}{N}i + 1 \right) \right\rceil.
\label{raw_optimal_non_singular_coding_equation}
\end{equation}
When $i$ is sufficiently large, we have  
\begin{equation}
l_i \approx \log_N \left(\frac{N-1}{N}i \right). 
\label{optimal_non_singular_coding_equation}
\end{equation}
Combining Eqs. \ref{optimal_coding_equation} and \ref{optimal_non_singular_coding_equation}, one obtains 
\begin{equation}
\log_N \frac{1}{p_i} \approx \log_N \left(\frac{N-1}{N}i \right) 
\end{equation}
and finally Zipf's law (Eq. \ref{Zipfs_law_equation}) with $\alpha = 1$. 

By presenting this derivation, we are not taking for granted that real language use is fully optimal with regard to any of the coding schemes mentioned above. Instead, our point is that it is not surprising that languages tend toward Zipf's law given the convenience of both kinds of compression for easy and fast communication \cite{Christiansen2015a}. 

Our derivation of Zipf's law is reminiscent of Mandelbrot's derivation based on typing at random on a keyboard (random typing) \cite{Mandelbrot1966} and defined by three parameters,  
$N$ (the alphabet size), $p_s$ (the probability of hitting the space bar) and $l_0$ (the minimum word length). The last parameter has been introduced to accommodate other variants of Mandelbrot's model \cite{Ferrer2009a}. Here we revisit his arguments from this generalized model and its predictions about word length. His central assumption is that typing at random determines the probability of a word, which has two key implications. First, a relationship between the length of a word and its probability \cite{Ferrer2015a} 
\begin{equation}
l = a \log_N p + b, \label{law_of_abbreviation_in_random_typing_equation}
\end{equation}
where $a$ and $b$ are constants ($a$ < 0) defined on the parameters of the model as 
\begin{equation}
a = \left( \log_N \frac{1-p_s}{N} \right)^{-1}
\end{equation} 
and 
\begin{equation}
b = a \log_N \frac{(1-p_s)^{l_0}}{p_s}.
\end{equation}
Second, a relationship between the length of a word and its rank that matches exactly that of optimal non-singular codes in Eq. \ref{raw_optimal_non_singular_coding_equation}.
Combining these two implications of random typing, one obtains 
\begin{equation}
\log_N \frac{N-1}{N}i \approx a \log_N p + b.
\end{equation}
With the assumption $b \approx 0$, one finally obtains Zipf's law for word frequencies with 
\begin{eqnarray}
\alpha & = & -1/a \\ 
       & = & \log_N \frac{N}{1-p_s} \\
       & = & 1 - \log_N (1 - p_s) \label{exponent_of_random_typing_equation} \\ 
       & > & 1.
\end{eqnarray}
Mandelbrot's derivation of Zipf's law is based on the assumption that $l_0 = 0$. While we start from an exact relationship between length and rank (Eq. \ref{raw_optimal_non_singular_coding_equation}) that we approximate later on, Mandelbrot presupposed that the relationship between length and rank {\em "cannot be represented by any simple analytical expression"} and started from an approximate equation for $l_0 = 0$ \cite{Mandelbrot1966}, i.e.
\begin{equation}
l \approx \log_N [(N-1)i],
\end{equation}
apparently unaware of the connection between random typing and optimal non-singular coding \cite{Mandelbrot1966}). Notice that Eq. \ref{law_of_abbreviation_in_random_typing_equation} can be interpreted, approximately, as a linear generalization of the relationship between $l$ and $p$ of optimal uniquely decipherable codes in Eq. \ref{optimal_coding_equation}. 
The exact (Eq. \ref{raw_optimal_non_singular_coding_equation}) and approximate (Eq. \ref{law_of_abbreviation_in_random_typing_equation}) connections between random typing and optimal coding challenge the view of random typing as totally detached from cost-cutting considerations \cite{Miller1963,Li1992b}.

Our derivation of Zipf's law presents various advantages over random typing. First, it starts from realistic cognitive pressures \cite{Christiansen2015a}.
Second, random typing is based exclusively on random choices but its parameters cannot be set at random: indeed, a precise tuning of the parameters is needed to mimic Zipf's law with $\alpha = 1$ \cite{Mandelbrot1966} (Eq. \ref{exponent_of_random_typing_equation} indicates that $\alpha$ depends on $N$ and $p_s$; the assumption $b\approx 0$ implies further parameter tuning). In contrast, our argument does not require parameter setting.  
Third, its assumptions are far reaching: compression allows one to shed light on the origins of three linguistic laws at the same time: Zipf's law for word frequencies, Zipf's law of abbreviation and Menzerath's law with the unifying principle of compression \cite{Ferrer2012d,Ferrer2015a,Gustison2016a}. There are many ways of explaining the origins of power-law-like distributions such as Zipf's law for word frequencies \cite{Stumpf2012a} but compression appears to be as the only one that can lead to a compact theory of statistical laws of language. 

Although uniquely decipherable codes are a subset of non-singular codes, it is tempting to think that both optimal non-singular coding and optimal uniquely decipherable coding cannot be satisfied to a large extent simultaneously. However, random typing is an example of how both constraints can be met approximately. We suggest that human languages are additional examples of a different nature. The two forms of optimality can coexist to some degree because the need for unique decipherability is alleviated by statistical cues that humans use to segment the linguistic input \cite{Saffran2003a}.

A challenge for the present derivation is that $\alpha$ is well-known to vary around 1 \cite{Alday2016a,Moreno2016a}. We speculate on some possibilities. One is that some variation of the exponent could be hidden by the assumptions of the model (e.g., that real languages are fully optimal) or by the approximations involved in our simple derivation of Zipf's law from compression (e.g., that ranks must be large enough). We consider a third possibility that could be purely formal.   
Suppose that a language uses an alphabet of $N$ symbols for optimal non-singular coding and an alphabet of $N'$ symbols for optimal uniquely decipherable coding. Suppose also that the lengths of a word in both schemes are comparable. 
Applying the arguments above with a suitable change of base of the logarithm one would get Zipf's law with 
\begin{equation}
\alpha = \log_N N'.  
\end{equation}
Thus $\alpha$ could vary above 1 if $N' > N$ or below 1 if $N' < N$. 

Here we have only sketched a new path to derive power-law-like distributions of ranks from efficiency considerations. We hope that our simple derivation 
and our speculations stimulate further research. 

\section*{Acknowledgements}

We thank two anonymous reviewers for their very valuable feedback.
We are also grateful to A. Corral, {\L}. D\k{e}bowski, S. Semple and C. Seguin for helpful comments and discussions. This research was funded by the grants 2014SGR 890 (MACDA) from AGAUR (Generalitat de Catalunya) and also
the APCOM project (TIN2014-57226-P) from MINECO (Ministerio de Economia y Competitividad).

\bibliographystyle{unsrt}

\bibliography{../law_of_abbreviation/biblio,../compression_theory/Ramon}

\end{document}